\documentclass[runningheads]{llncs}
\usepackage[T1]{fontenc}
\usepackage{graphicx}
\def\Args{Args}
\def\Att{Att}

\def\wrt{wrt}

\newcommand{\abaf}{\ensuremath{\langle {\cal L}, \, {\cal R}, \, {\cal A},\, \overline{ \vrule height 5pt depth 3.5pt width 0pt \hskip0.5em\kern0.4em}\rangle}}
\def\contrary{\overline{ \vrule height 5pt depth 3.5pt width 0pt \hskip0.5em\kern0.4em}}

\newcommand{\abafpone}{\ensuremath{\langle {\cal L}, \, {\cal R}', \, {\cal A},\, \overline{ \vrule height 5pt depth 3.5pt width 0pt \hskip0.5em\kern0.4em}\rangle}}

\newcommand{\abafp}{\ensuremath{\langle {\cal L}', \, {\cal R}', \, {\cal A}',\, \overline{ \vrule height 5pt depth 3.5pt width 0pt \hskip0.5em\kern0.4em}'\rangle}}

\newcommand{\lang}{\ensuremath{\mathcal{L}}}
\newcommand{\asms}{\ensuremath{\mathcal{A}}}
\newcommand{\rules}{\ensuremath{\mathcal{R}}}

\newcommand{\argur}[3]{#1 \vdash_{#3} #2} 
\newcommand{\argu}[2]{#1 \vdash #2} 
\newcommand{\ruleset}{\ensuremath{R}}
\newcommand{\sent}{\ensuremath{s}}

\newcommand{\asm}{\ensuremath{a}}
\newcommand{\asmset}{\ensuremath{A}}

\newcommand{\calpha}{c\mbox{-}\alpha}
\newcommand{\flies}{\mathit{flies}}

\usepackage{algorithm}
\usepackage{algorithmic}
\usepackage{amsmath}
\usepackage{amssymb}
\usepackage{graphicx}
\usepackage[colorlinks=true, allcolors=blue]{hyperref}
\usepackage{xcolor}
\usepackage{cancel}

\newtheorem{conj}{Property}

\newcommand{\pick}{choose}

\newcommand{\A}{\mathcal{A}}
\newcommand{\LL}{\mathcal{L}}
\newcommand{\RR}{\mathcal{R}}
\newcommand{\ABA}{\langle \LL,\RR,\A,\contrary\rangle}

\newcommand{\sABA}{\langle\RR,\A,\contrary\rangle}
\newcommand{\sABAp}{\langle \RR',\A',\contrary'\rangle}

\newcommand{\vdashunacc}{\ensuremath{\not \models}}
\newcommand{\vdashacc}{\ensuremath{\models}}

\newcommand{\mpnote}[1]{\noindent \textcolor{blue!60}{(Maurizio)} \textcolor{blue!90}{#1}}
\newcommand{\ftnote}[1]{\noindent \textcolor{orange!100}{(Francesca)} \textcolor{orange!100}{#1}}
\newcommand{\mprem}[1]{\noindent \textcolor{gray!100}{\%} \textcolor{gray!100}{#1}}
\newcommand{\rem}[1]{\noindent \textcolor{green!100}{\%} \textcolor{green!100}{#1}}

\begin{document}

\title{Learning Assumption-based Argumentation Frameworks\thanks{ILP 2022, 31st International Conference on Inductive Logic Programming, Cumberland Lodge, Windsor, UK}
}
\titlerunning{Learning ABA Frameworks}
%
\author{Maurizio Proietti \inst{1}\orcidID{0000-0003-3835-4931} 
\and
Francesca Toni\inst{2}\orcidID{0000-0001-8194-1459} }
\authorrunning{M. Proietti and F. Toni}
%
\institute{IASI-CNR, Rome, Italy \\
\email{maurizio.proietti@iasi.cnr.it}
\and
Imperial College London, UK\\
\email{ft@ic.ac.uk}}
\maketitle              

\vspace*{-6mm}

\begin{abstract}
We propose a novel approach to logic-based learning which generates \emph{assumption-based argumentation (ABA) frameworks}
from positive and negative examples, using a given background knowledge. 
These ABA frameworks can be mapped onto logic programs with negation as failure that may be non-stratified. 
Whereas existing argumentation-based methods 
learn exceptions to general rules by interpreting the exceptions as \emph{rebuttal attacks}, our approach interprets them as \emph{undercutting attacks}.
Our learning technique is based on the use of transformation rules, including some adapted from logic program transformation rules (notably \emph{folding}) as well as others, such as \emph{rote learning} and \emph{assumption introduction}. 
We present a general strategy that applies the transformation rules
in a suitable order to learn stratified frameworks, and we also propose a variant that handles the non-stratified case. 
We illustrate the benefits of our approach with a number of examples,
which show that, on one hand, we are able to easily reconstruct other logic-based learning approaches and, on the other hand, we can work out in a very simple and natural way problems that seem to be hard for existing techniques.

\vspace*{-3mm}
\keywords{Logic-based learning  \and Assumption-based argumentation \and Logic program transformation.}
\end{abstract}

\vspace*{-6mm}

\section{Introduction}

\vspace*{-2mm}

Various forms of \emph{computational argumentation}, notably abstract argumentation \cite{Dung_95} and assumption-based argumentation (ABA) \cite{ABA,ABAbook,ABAtutorial} and \cite{ABAhandbook}, have been advocated as unifying frameworks for various forms of non-monotonic reasoning, including logic programming. 
However,
with very few early exceptions (notably \cite{DimopoulosK95}), 
computational argumentation has received little attention as a basis to support logic-based learning. 
Here, we fill this gap by proposing a novel 
approach
 to logic-based learning which
generates ABA frameworks from 
examples, using background knowledge also in ABA format. 

In general, 
ABA frameworks amount to a \emph{deductive system} (consisting of a logical \emph{language} to specify sentences and of a set of inference \emph{rules} built from these sentences), a set of \emph{assumptions}   (which are sentences with a special status) and their \emph{contraries} (which are sentences).
Rules are then used to construct \emph{arguments}, which support claims by means of assumptions, and  \emph{attacks} between assumptions/arguments are achieved by means of arguments supporting contraries of assumptions. 
The semantics of ABA frameworks is then defined in terms of various notions of  \emph{(dialectically) acceptable extensions}, amounting to sets of assumptions/arguments that can be deemed to defend themselves against attacks, in some sense (see Sect.~\ref{sec:ABA}). 
Thus, the goal of  \emph{ABA learning} amounts to identifying rules, assumptions and their contraries (adding to those in the given background knowledge) that cover argumentatively the positive examples and none of the negative examples 
according to some chosen semantics (see Sect.~\ref{sec:goal}).   

Logic programs can be seen as restricted instances of ABA frameworks  (
where
assumptions are negation as failure literals, and the contrary of $not \, p$ is $p$) \cite{ABA}. Moreover, ABA frameworks of a special kind (namely when all sentences are atomic and no assumption occurs in the head of rules) can be mapped to logic programs. These mappings hold under a range of semantics for ABA and for logic programming \cite{ABA}.  Thus, learning ABA  frameworks can be seen as learning logic programs. However, given that ABA frameworks admit several other non-monotonic reasoning instances \cite{ABA}, learning ABA frameworks can also in principle provide a way to learn other forms of non-monotonic logics. 

Our approach to learning ABA frameworks relies upon  transformation rules applied to ABA frameworks (to obtain new ones, see Sect.~\ref{sec:transformation}): some of these transformation rules are adapted from logic programming, such as \emph{folding} \cite{PeP94}, whereas others are new, such as \emph{rote learning} and \emph{assumption
introduction}. We define the transformation rules and outline a general strategy for applying them (see Sect.~\ref{sec:strategy}) in order to learn ABA frameworks that achieve the goal of ABA learning. We also illustrate, with the use of examples (see Sect.~\ref{sec:undercutting},~\ref{sec:nixon}), the benefits of ABA learning in comparison with existing methods. We focus on similarly inspired methods, notably \cite{DimopoulosK95} and \cite{ShakerinSG17}: the former learns argumentation frameworks of a different kind (where arguments are built from clauses and preferences over them), whereas the latter learns logic programs with negation as failure, but can be seen as being guided by argumentative principles (while not being explicitly so).  
Both these methods learn exceptions to general rules by interpreting the exceptions as \emph{rebuttal attacks} (namely attacks between arguments with conflicting claims), and result in stratified logic programs~\cite{Ap&88}. Instead, our approach interprets exceptions as 
\emph{undercutting attacks} (from an argument to the support of another) while also accommodating rebuttal attacks, and may give rise to  
logic programs with non-stratified negation
as failure. 

\paragraph{Related work.} Several approaches to 
learning logic programs with negation as failure exist, in addition to the aforementioned \cite{DimopoulosK95,ShakerinSG17}.
These include
{methods for learning exceptions to default rules \cite{InoueK97}},
Sakama's 
inverse entailment~\cite{Sakama00}
and brave induction ~\cite{SakamaI09} in non-monotonic logic programs,
and induction from answer sets~\cite{Sakama05}. Answer sets are also the target of other logic-learning methods, notably as in ILASP ~\cite{LawRB14}.
These approaches 
learn non-stratified logic programs, 
{rather than ABA frameworks}. 
ABA can be seen as performing abductive reasoning (in that assumptions are hypotheses open for debate).  
Whereas other approaches combine  
abductive  and inductive learning~\cite{Ray09}, 
again, they 
{do not learn ABA frameworks}.  
{Some approaches learn abductive logic programs~\cite{InoueH00}, which rely upon assumptions, like ABA frameworks. Overall, providing a formal comparison between our method and existing methods learning instances of ABA frameworks is left for future work. Finally, several existing works use argumentation to perform learning (e.g., see overview in \cite{ML-argsurvey}) while others learn argumentation frameworks (e.g., recently, \cite{dear}): we differ from the former in that our method learns ABA frameworks rather than using argumentation to aid learning, and from the latter  in that we learn a different type of argumentation frameworks. }

\section{Background: Assumption-based argumentation (ABA)} 
\label{sec:ABA}

An {\em Assumption-based argumentation (ABA) framework} (as originally proposed in \cite{ABA}, but presented here following  recent accounts in  \cite{ABAbook,ABAtutorial} and \cite{ABAhandbook}) is a tuple \abaf{}
where

\vspace*{-0.2cm}
\begin{itemize}
\item 
$\langle \lang, \rules\rangle$ is a deductive system,
 where $\lang$ is a \emph{language} and $\rules$ is a set of
 \emph{(inference) rules} of the form $\sent_0 \leftarrow \sent_1,\ldots, \sent_m $ ($m \ge 0, \sent_i \in \lang$, for $1\leq i \leq m$); 

\item 
$\asms$ $\subseteq$ $\lang$ is a (non-empty) 
set
of {\em assumptions};\footnote{The non-emptiness requirement
 can always be satisfied by including in $\A$ a \emph{bogus assumption}, with its own contrary, neither occurring elsewhere in the ABA framework. 
For 
conciseness, we will not write this assumption and its contrary explicitly.} 

\item 
$\contrary$ is a total mapping from $\asms$ into
 $\lang$, where $\overline{\asm}$ is the {\em contrary} of $\asm$, for $\asm
  \in \asms$.
\end{itemize}
\vspace*{-0.2cm}
Given a rule $\sent_0 \gets \sent_1, \ldots,
\sent_m$, $\sent_0$ is 
the {\em head} 
and $\sent_1,\ldots, \sent_m$ 
is 
the {\em body}; 
if $m=0$ then the  body is said to be {\em empty} (represented as $\sent_0 \gets$ or  $\sent_0 \gets true$).
If assumptions are not heads 
of rules then the ABA framework is called
{\em flat}. 

Elements of $\LL$ can be any sentences, but in this paper we will focus on (flat) ABA frameworks where $\LL$ is a set of ground atoms.
However, in the spirit of logic programming, we will use \emph{schemata} for rules, assumptions and contraries, using variables to represent compactly all instances over some underlying universe. 

\begin{example}
\label{ex:simple}
Let $\LL=\{p(X), q(X), r(X), a(X), b(X) | X\in \{1,2\}\}$. 
Then, 
an ABA framework may be \abaf{}
 with the other components defined as follows:

 $\RR=\{p(X) \gets a(X), \; \;\; q(X)\gets b(X), \;\; \; r(1) \gets true\}$;
    
    $\A=\{a(X), b(X)
    \}$; 
    
    $\overline{a(X)}=q(X)$, \quad $\overline{b(X)}=r(X)$.
\end{example}
\vspace*{-0.2cm}
In ABA, {\em arguments} are deductions of claims using rules and 
supported by assumptions, and {\em attacks} are directed at the
assumptions in the support of arguments.  
More formally, following
\cite{ABAbook,ABAtutorial,ABAhandbook}:
\vspace*{-0.2cm}
\begin{itemize}
\item 
\emph{An argument for (the claim) $\sent \in \lang$ 
supported by $\asmset \subseteq \asms$ and $\ruleset \subseteq \rules$
}
(denoted $\argur{\asmset}{\sent}{\ruleset}$
) is a finite tree with nodes labelled by
sentences in $\lang$ or by $true$
, the root labelled by $\sent$, leaves either $true$ or
assumptions in $\asmset$, and non-leaves $\sent'$ with, as children,
the elements of the body of some rule in \ruleset{} 
with head $\sent'$.
\item 
 Argument 
$\argur{\asmset_1}{\sent_1}{\ruleset_1}$ 
{\em attacks} argument
$\argur{\asmset_2}{\sent_2}{\ruleset_2}$  
iff
$\sent_1=\overline{\asm}$ for some
$\asm \in \asmset_2$.
\end{itemize}
\vspace*{-0.2cm}
\begin{example}[Ex.~\ref{ex:simple} cntd]
\label{ex:simple-AA}
Let rules in $\RR$ be named $\rho_1(X)$, $\rho_2(X)$, $\rho_3$, respectively. 
Then, arguments include 
$\argur{\{b(1)\}}{q(1)}{\{\rho_2(1)\}}$,
$\argur{\emptyset}{r(1)}{\{\rho_3\}}$, and 
$\argur{\{a(1)\}}{a(1)}{\emptyset}$,\footnote{The other (ground) arguments are 
$\argur{\{a(1)\}}{p(1)}{\{\rho_1(1)\}}$,
$\argur{\{a(2)\}}{p(2)}{\{\rho_1(2)\}}$,
$\argur{\{b(2)\}}{q(2)}{\{\rho_2(2)\}}$,
$\argur{\{a(2)\}}{a(2)}{\emptyset}$,
$\argur{\{b(1)\}}{b(1)}{\emptyset}$, and
$\argur{\{b(2)\}}{b(2)}{\emptyset}$.}
where, for example, $\argur{\emptyset}{r(1)}{\{\rho_3\}}$ attacks
$\argur{\{b(1)\}}{q(1)}{\{\rho_2(1)\}}$.
\end{example}

Given argument $\alpha=\argur{\asmset}{\sent}{\ruleset}$, we will refer to the single rule $\rho \in R$ such that the sentences in $\rho$'s body label $\sent$' children  
in $\alpha$ as the \emph{top rule} of the argument. 

Note that, in ABA, attacks are   carried out by \emph{undercutting} some of their supporting premises (the assumptions). Other forms of attacks can be also represented in ABA
, including attacks by \emph{rebuttal}, where arguments disagree on their claim  (see \cite{ABAtutorial} for details).    

Given a flat ABA framework \abaf, let $\Args$ be the set of all arguments
and $\Att=\{(\alpha,\beta) \in \Args \times \Args \mid \alpha$ attacks $\beta\}$, for `arguments' and `attacks' defined as above.
Then $(\Args,\Att)$ is an Abstract Argumentation (AA) framework \cite{Dung_95} and standard semantics 
for the latter 
can be used to determine  the semantics of the ABA framework
\cite{ABAtutorial}.\footnote{ABA semantics were originally defined in terms of sets of assumptions and attacks between them \cite{ABA}, but can be reformulated, for flat ABA frameworks, 
in terms of sets of arguments and attacks between them (see~\cite{ABAtutorial}), as given here. }
For example, $S\subseteq \Args$ is a \emph{stable extension} iff (i) $\nexists \alpha,\beta \!\in \!S$ such that $(\alpha,\beta) \!\in \!\Att$ (i.e. $S$ is \emph{conflict-free}) and (ii) $\forall \beta \!\in \!\Args\setminus S, \exists \alpha \!\in \!S$ such that $(\alpha,\beta) \!\in \!\Att$ (i.e. $S$ ``attacks'' all arguments it does not contain).

In the following example and in the remainder we will omit to specify the supporting rules in arguments (because these rules play no role when determining extensions
, as attacks only depend on supporting assumptions and claims
). Thus, 
e.g., argument $\argur{\{b(1)\}}{q(1)}{\{\rho_2(1)\}}$ in Ex.~\ref{ex:simple-AA} is written simply as $\argu{\{b(1)\}}{q(1)}$.

\begin{example}[Ex.~\ref{ex:simple-AA} cntd]
\label{ex:simple-sem}
The AA framework is $(\Args,\Att)$ with $\Args$ as given earlier and $\Att=\{
(\argu{\emptyset}{\!r(1)},\argu{\{b(1)\}}{q(1)}), (\argu{\emptyset}{r(1)},\argu{\{b(1)\}}{b(1)})\} \cup \{(\argu{\{b(X)\}}{q(X)},\argu{\{a(X)\}}{p(X)}), (\argu{\{b(X)\}}{q(X)},\argu{\{a(X)\}}{a(X)}) | X \in \{1,2\}\}$. Then,
$\{\argu{\emptyset}{r(1)}, \argu{\{a(1)\}}{p(1)}, \argu{\{a(1)\}}{a(1)}, \argu{\{b(2)\}}{q(2)}, \argu{\{b(2)\}}{b(2)}\}$ is the only stable extension. This captures the ``model'' $\{r(1), p(1), a(1), q(2), b(2)\}$
, amounting to all the claims of accepted arguments in the stable extension~\cite{ABAtutorial}. 
\end{example}
\vspace*{-0.1cm}
Note that, in general, ABA frameworks may admit several stable extensions, in which case one can reason \emph{credulously} or \emph{sceptically},
by focusing respectively on (claims of arguments accepted in) a single 
or all stable extensions.  

\begin{example}
\label{ex:simple-cred}
Given \abaf{} as in Ex.~\ref{ex:simple},
consider 
\abafpone{}, where $\RR'=\RR \cup\{r(2) \gets a(2)\}$. Now, there are two stable extensions: the same as in Ex.~\ref{ex:simple-sem} as well as 
$\{\argu{\emptyset}{r(1)}, \argu{\{a(1)\}}{p(1)}, \argu{\{a(1)\}}{a(1)}, \argu{\{a(2)\}}{p(2)}, \argu{\{a(2)\}}{a(2)}, \argu{\{a(2)\}}{r(2)}\}$, with accepted claims $\{r(1), a(1),$ $ p(1), a(2), p(2), r(2)\}$. Then, e.g., $r(1)$ is sceptically accepted whereas $r(2)$ is only credulously accepted.
\end{example}
\vspace*{-0.1cm}
Finally, note that ABA frameworks of the form considered here can be naturally  mapped onto logic programs where assumptions are replaced by the negation as failure of their contraries, such that, in particular, stable extensions of the former correspond to stable models \cite{SM} of the latter.\footnote{The correspondence also holds under other semantics, omitted here for simplicity.}
\begin{example}[Ex.~\ref{ex:simple-cred} cntd]
For instance, \abafpone\ from Ex.~\ref{ex:simple-cred} can be mapped onto the logic program\footnote{We use the same notation for ABA rules and logic programs as, indeed, logic programming is an instance of ABA~\cite{ABA}.}

$p(X)\gets not\,q(X), \quad  q(X)\gets not \, r(X), 
\quad r(1) \gets, 
\quad \;\, r(2) \gets not \,q(2).
$

\smallskip

\noindent This admits two stable models, identical to the claims accepted in the two stable extensions for the ABA framework.
\end{example}
Thus, learning ABA frameworks of the form we consider amounts to a form of inductive logic programming (ILP).

In the remainder,  without loss of generality we leave the language component of all ABA frameworks implicit,
and use, e.g., $\sABA $  to stand for $\ABA $ where $\LL$ is the set of all sentences in $\RR$, $\A$ and in the range of $\contrary$.
We will also use $\sABA \vdashacc s$  to indicate 
that $s\in \LL$ is the claim of an argument accepted in all or some stable extensions (depending on the choice of reasoning) of $\sABA$. 

\section{The ABA learning problem}
\label{sec:goal}

We define the core ingredients of ABA learning, in a similar vein as in ILP.

\begin{definition}
A \emph{positive/negative example (for a predicate $p$ with arity $n \geq 0$)} is a ground atom of the form $p(c)$, for $c$ a tuple of $n$ constants.
\end{definition}
The predicate amounts to the \emph{concept} that inductive ABA aims to learn (in the form of an ABA framework whose language component includes the examples amongst its sentences).  Positive and negative examples, albeit having the same syntax, play a very different role in ABA learning, as expected.
Like standard ILP,  ABA learning is driven by a (\emph{non-empty}) set of positive examples and a (\emph{possibly empty}) set of negative examples, with the help of some \emph{background knowledge} compactly providing information about the examples, as follows: 

\begin{definition}
\label{def:back}
The \emph{background knowledge} 
is 
an  ABA framework
. 
\end{definition}
%
An illustration of possible background knowledge  $\sABA$ and examples for a simple inductive ABA task is shown in Fig.~\ref{fig:penguin} (adapted from \cite{DimopoulosK95}). 
\begin{figure}[t]
\begin{eqnarray*}
& \RR = &\{bird(X) \gets penguin(X), \quad penguin(X) \leftarrow superpenguin(X),  \\ 
&& bird(a)\gets, \;
bird(b)\gets, \quad
 penguin(c)\gets,\; 
penguin(d)\gets, \\
&& superpenguin(e)\gets, \; 
superpenguin(f)\gets\} 
\end{eqnarray*}
Positive Examples:
\(
\{\flies(a),
\flies(b),
\flies(e),
\flies(f)\}
\)

Negative Examples:
\(\{\flies(c),
\flies(d)
\}\)
\caption{
A simple illustration of  background knowledge as an ABA framework $\sABA$ (showing only $\RR$, with $\A$ consisting of a single bogus assumption), and of positive/negative examples  for ABA learning of concept $\flies/1$.}
\label{fig:penguin}
\vspace{-5mm}
\end{figure}
Then, in  ABA learning, we establish whether examples are \emph{(not) covered} 
argumentatively:

\begin{definition}
 Given an ABA framework $\sABA$, an example $e$ is \emph{covered} by  $\sABA$ iff  $\sABA \vdashacc e$ and is \emph{not covered} by  $\sABA$ iff $\sABA \vdashunacc e$.
\end{definition}
In the case of Fig.~\ref{fig:penguin},
none of the examples is covered by the background knowledge 
(no matter 
what \vdashacc\ is,
as no arguments for any of the examples exist). 

\begin{definition}
Given background knowledge $\sABA$, positive examples $\mathcal{E}^+$ and
negative examples $\mathcal{E}^-$,
with $\mathcal{E}^+\cap \mathcal{E}^- =\emptyset$,
the \emph{goal of ABA learning} is to 
construct
$\sABAp$ such that 
$\RR \subseteq \RR'$, 
$\A \subseteq \A'$,
and, for all $\alpha \in \A$, $\overline{\alpha}'=\overline{\alpha}$, such that:

\noindent 
(Existence) $\sABAp$ admits at least one extension under the chosen ABA
semantics,

\noindent
(Completeness) for all $e \in \mathcal{E}^+$, $\sABAp \vdashacc e$, and

\noindent
(Consistency) for all $e \in \mathcal{E}^-$, $\sABAp \vdashunacc e$.

\end{definition}
%
This definition is parametric \wrt\ the choice of semantics, and also whether this is used credulously or sceptically. 
For choices of semantics other than stable extensions (e.g., grounded or preferred 
extensions~\cite{ABAhandbook})
the Existence requirement may be trivially satisfied (as these extensions 
always 
exist).

\begin{figure}[t]
\begin{eqnarray*}
&& \RR' =  \{\flies(X) \gets bird(X),\alpha_1(X), \\
&& \quad\quad\quad \calpha_1(X) \gets penguin(X),
\alpha_2(X),\\ && \quad\quad
\quad c\mbox{-}\alpha_2(X) \gets superpenguin(X)
\} \cup \RR 
\\
&& \A'= \{\alpha_1(X), \alpha_2(X)\} 
\quad \text{ with } \overline{\alpha_1(X)}'=\calpha_1(X), \quad \overline{\alpha_2(X)}'=\calpha_2(X)
\end{eqnarray*}
\caption{
An ABA framework $\sABAp$ achieving the goal  of ABA learning, given the background knowledge and examples in Fig.~\ref{fig:penguin}.}
\label{fig:penguin-learnt}
\end{figure}
The ABA framework $\sABAp$ in Fig.~\ref{fig:penguin-learnt} 
admits a single stable extension and  covers all positive examples and none of the negative examples in Fig.~\ref{fig:penguin} (for either the sceptical or credulous version of $\vdashacc$
). Thus, this $\sABAp$ achieves the goal of ABA learning. 
But how can this ABA framework be obtained? 
In the remainder 
we will outline our method for achieving the goal of ABA learning.

\section{Transformation rules for ABA frameworks}
\label{sec:transformation}

Our method for solving the ABA learning problem makes use of transformation rules
, some of which are borrowed from the field of logic program transformation~\cite{ArD95,PeP94,Sek91,ToK96}, while
others are specific for ABA frameworks.


We assume that, for any ABA framework, the language $\LL$ contains all equalities between elements of the underlying universe and $\mathcal R$ includes all rules $a=a\leftarrow$, where $a$ is an element of the universe.
We also assume that all rules in $\mathcal{R}$ (except for the implicit equality rules) are \emph{normalised}, i.e. they are written as:

\smallskip

$p_0({X}_0) \leftarrow eq_1, \ldots, eq_k, p_1({X}_1), \ldots, p_n({X}_n)$

\smallskip
\noindent
where $p_i({X}_i)$, for $0\leq i \leq n$,  is an atom (whose ground instances are) in $\mathcal L$ and $eq_i$, for $1\leq i \leq k$, is an equality whose variables occur in the tuples ${X}_0,{X}_1, \ldots, {X}_n$.
The body of a normalised rule can be freely rewritten by using the standard axioms of equality, e.g., $Y_1=a, Y_2=a$ can be rewritten as $Y_1=Y_2, Y_2=a$.
\smallskip

Given an ABA framework $\sABA$, a {\em transformation rule} 
constructs a new ABA framework $\sABAp$. 
We use 
transformation rules R1--R5 below,
where we mention explicitly only the components of the 
framework that are modified.

\smallskip

\noindent
\emph{R1. Rote Learning.} Given 
atom $p({t})$,
add 
%
%
$\rho :  p({X}) \leftarrow  {X}={t}$ 
%
to $\RR$. Thus, $\RR'\!=\!\RR\cup \{\rho\}$.

\smallskip

In our learning strategy, Rote Learning 
is typically applied with $p({t})$ a positive example. The added rule  allows us 
to obtain an argument for $p({t})$.

\begin{example}
\label{ex:R1}
Let us consider the learning problem of Fig.~\ref{fig:penguin}.
By four applications of 
R1,
from the positive examples $\flies(a),\flies(b),\flies(e),\flies(f)$,
we get

$\RR_1 =   \RR \cup \{\flies(X) \gets X=a, 
\flies(X) \gets X=b,$ 

~~~~~~~~~~~~~~~$\flies(X) \gets X=e, 
\flies(X) \gets X=f\}$.
\end{example}
\smallskip

\noindent
\emph{R2. Equality Removal.} Replace a rule 
%
$\rho_1: H \leftarrow eq_1, Eqs, B$ in $\RR$,
%
where $eq_1$ is an equality, $Eqs$ is a (possibly empty) set of equalities, and $B$ is a (possibly empty) set of atoms,
by 
rule
%
$\rho_2: H \leftarrow Eqs, B.$
%
Thus, $\RR'=(\RR\setminus \{C_1\}) \cup \{C_2\}$.


\smallskip

Equality Removal allows us to generalise a rule by deleting an equality in its body. We will see an example of application of this rule in Sect.~\ref{sec:undercutting}. 
\smallskip

\noindent
\emph{R3. Folding.}
Given rules
%

$\rho_1$: $H \leftarrow Eqs_1, B_1, B_2$
\quad and \quad
$\rho_2$: $K \leftarrow Eqs_1, Eqs_2, B_1$

\smallskip

\noindent
in $\RR$, where $\textit{vars}(\textit{Eqs}_2) \cap \textit{vars}((H,B_2))=\emptyset$, replace $\rho_1$ by 
%
$\rho_3$: $H \leftarrow Eqs_2, K, B_2.$
%
%
Thus, $\RR'=(\RR\setminus \{\rho_1\})\cup\{\rho_3\}$.

\smallskip

Folding is a form of {\em inverse resolution}~\cite{Muggleton1995}, 
used for generalising a rule by replacing 
some atoms in its body with their `consequence' using a rule in $\RR$.

\begin{example}
\label{ex:R3}
By 
R3 using 
rule $bird(X) \gets X=a$ in $\RR$, rule $\flies(X) \gets X=a$ in $\RR_1$ (see Ex.~\ref{ex:R1}) is replaced by
$\flies(X) \gets bird(X)$, 
giving 
$\RR_2 \!=\!   \RR \cup \{\flies(X) \gets bird(X),\
\flies(X) \gets X=b, \flies(X) \gets X=e,\
\flies(X) \gets X=f\}.$
\end{example}

\smallskip

\noindent
\emph{R4. Subsumption.}
Suppose that  $\RR$ 
contains rules

$\rho_1: H \leftarrow Eqs_1, B_1$
\quad and \quad
$\rho_2: H \leftarrow Eqs_2, B_2$

\noindent
such that, for every ground instance $H' \leftarrow Eqs'_2, B'_2$ of $\rho_2$,
there exists a ground instance $H' \leftarrow Eqs'_1, B'_1$  of $\rho_1$ (with the same head)
such that, if for each atom $B$ in $Eqs'_2, B'_2$ there is an argument $\argur{S}{B}{R}$,
then for each atom $A$ in $Eqs'_1, B'_1$ there is an argument $\argur{S'}{A}{R'}$ 
with $S'\subseteq S$.
Thus, $\rho_2$ is 
\emph{subsumed} by $\rho_1$ and can be deleted from $\RR$, and hence $\RR'=\RR\setminus \{\rho_2\}$.


In particular, R4 generalises the usual logic program transformation that allows the removal of a rule which is
$\theta$-subsumed by another one~\cite{PeP94}.

\begin{example}
Rule $\flies(X) \gets X\!=\!b$ in $\RR_2$ (from Ex.\ref{ex:R3}) is subsumed by 
$\flies(X) \gets bird(X)$. Indeed, $b$ is the only ground $x$ such that
$\argur{\emptyset}{x\!=\!b}{\{x=b \gets \}}$,
and when $x$ is  $b$, we have $\argur{\emptyset}{bird(x)}{\{bird(x) \gets x=b,\,  x=b \gets \}}$.
Similarly, rules $\flies(X) \gets X\!=\!e$ and $\flies(X) \gets X\!=\!f$ 
are subsumed by $\flies(X) \gets bird(X).$
Thus, we derive 
$\RR_3 = \RR \cup \{\flies(X) \gets bird(X)\}$.
\end{example}

\smallskip

\noindent \emph{R5. Assumption Introduction.} 
Replace 
$\rho_1: H \leftarrow Eqs, B$
%
%
in $\RR$ 
by
$\rho_2: H \leftarrow Eqs, B, \alpha({X})$
%
where $X$ is a tuple of variables taken from $\textit{vars}(H) \cup \textit{vars}(B)$ and $\alpha({X})$ is a (possibly new)
assumption with contrary $\chi({X})$. 
Thus, $\RR'=(\RR\setminus \{\rho_1\})\cup\{\rho_2\}$, $\mathcal{A}'= \mathcal{A} \cup \{\alpha({X})\}$, $\overline{\alpha({X})}'=\chi({X})$,
and $\overline{\beta}'=\overline{\beta}$ for all $\beta \in \A$.

\begin{example}
By applying R5, we 
get 
$\RR_4 \!\!=\!\! \RR \!\cup \{\flies(X) \!\!\gets\!\! bird(X),\alpha_1(X)\}$, 
with $\overline{\alpha_1(X)}'\!\!=\!\calpha_1(X)$.
%
Now, we complete this example to show, informally, that other approaches based on learning exceptions~\cite{DimopoulosK95,ShakerinSG17} 
can be recast in our setting. For the newly
introduced 
$\calpha_1(X)$,
we have positive examples $\mathcal E^+_1 =\{\calpha_1(c),\ \calpha_1(d)\}$, 
corresponding to the exceptions to $\flies(X) \gets bird(X)$,
and negative examples $\mathcal E^-_1 = \{\calpha_1(a),\calpha_1(b),\calpha_1(e),\calpha_1(f)\}$,
corresponding to the positive examples for $\flies(X)$ 
to which $\flies(X) \gets bird(X)$ correctly applies.
Then, starting from these examples, we can learn rules for
$\calpha_1(X)$ similarly to what we have done for $\flies$.
By Rote Learning, we get $\RR_5 = \RR \cup \{\flies(X) \gets bird(X),\alpha_1(X),\ \calpha_1(X) \gets X=c,\ \calpha_1(X)\gets X=d\}$. 
The rules for $\calpha_1(X)$ can be viewed as
rebuttal attacks 
against $\flies(X) \gets bird(X)$, as
$c$ and $d$ are birds that do not fly.
By Folding and Subsumption, we generalise the rules for 
$\calpha_1(X)$ and obtain $\RR_6 = \RR \cup \{\flies(X) \gets bird(X),\alpha_1(X),\ \calpha_1(X) \gets penguin(X)\}$. 
The new rule also covers the negative examples 
$\calpha_1(e)$ and $\calpha_1(f)$, and hence we 
introduce a new assumption $\alpha_2(X)$ with contrary
$\calpha_2(X)$ having $\calpha_2(e)$ and $\calpha_2(f)$
as positive examples.
By one more sequence of applications of Rote Learning,
Folding and Subsumption, we get exactly 
the 
ABA framework in Fig.~\ref{fig:penguin-learnt}.
\end{example}

In the field of logic program transformation, 
the goal is to derive new programs 
that
are \emph{equivalent},
\wrt\ a semantics of choice, to the initial program.
Various results guarantee that, 
under suitable conditions, transformation rules such
as Unfolding and Folding indeed enforce equivalence 
(e.g., \wrt\ the least Herbrand model of definite  programs~\cite{PeP94} 
and the stable model semantics of normal logic programs~\cite{ArD95}).
These results have also been generalised by using argumentative notions~\cite{ToK96}.

In the context of  ABA learning, however, program
equivalence is not a desirable objective, as we look for
sets of rules that generalise positive examples and avoid
to cover negative examples.
In particular, as shown by the examples of this and
next section, 
the generalisation of positive examples is done by
applying the Folding, Equality Removal, and Subsumption 
transformations, while the exceptions due to negative 
examples are learned by Assumption Introduction and Rote Learning.
A general strategy for applying the transformation
rules will be presented in Sect.~\ref{sec:strategy}.
The requirements that R2--R4 
should preserve positive examples, while
R1 and R5 should be able to
avoid negative examples, are formalised by the following two
properties that we require to hold. 
These properties are sufficient 
to show that, when the learning process terminates, it
indeed gets a solution of the ABA learning problem given in
input.

\begin{conj}\label{conj:gen}
Let $\sABAp$ be obtained by applying any of 
Folding, Equality Removal and Subsumption to $\sABA$ to modify rules with $p$ in the head
. 
If $\sABA \vdashacc p(t)$ then $\sABAp \vdashacc p(t)$. 
\end{conj}
\vspace*{-0.2cm}

\begin{conj}\label{conj:exception}
Let  $p(t_1), p(t_2)$ be atoms such that $p(t_1) \neq p(t_2)$ and $\sABA$ be such that $\sABA \vdashacc p(t_1)$ and $\sABA \vdashacc p(t_2)$.  
Then there exists
$\sABAp$ obtained from $\sABA$ by applying 
 Assumption Introduction to modify rules with $p$ in the head and {then applying Rote Learning to add rules for the
contraries of the assumption atoms},
 such that 
$\sABAp \vdashacc p(t_1)$ and $\sABAp \vdashunacc p(t_2)$.
\end{conj}


We leave 
to future work the identification
of 
conditions that enforce Properties~\ref{conj:gen},~\ref{conj:exception} 
\wrt~ABA semantics
.
However, note that, 
even though we cannot directly apply the results
about logic program transformations to prove 
these properties
, 
we can leverage the proof
methodology and adapt some partial results, and specifically
the ones developed 
by using argumentative notions~\cite{ToK96}.

\section{Learning by rebuttal and undercutting attacks}
\label{sec:undercutting}

Some approaches~\cite{DimopoulosK95,ShakerinSG17} propose learning techniques
for non-monotonic logic programs based on the idea of learning 
default rules with exceptions.
Default rules are learned by generalising positive examples,
and exceptions to those rules are learned by generalising negative examples.
This process can be iterated 
by learning exceptions to the 
exceptions
.
From an 
argumentative perspective, these approaches can be 
viewed as learning rules that attack each other by rebuttal,
because their heads are 
`opposite'.
We have illustrated through the $\flies$ example in Sect.~\ref{sec:transformation} that 
our transformation rules can easily reconstruct  learning by rebuttal.
Through another example, in this section we show that our transformation rules can also learn undercutting attacks, which in contrast 
seem hard for other approaches that learn by rebuttal.

Suppose that a robot moves through locations $1,\ldots,6$. Let atom $step(X,Y)$ mean that the robot can take a step from location $X$ to location $Y$. Some locations are busy, and the robot cannot 
occupy them. 
Consider the background knowledge with $\RR$ as follows (and a single bogus assumption): 


$\mathcal R = \{
step(1,2)\leftarrow,
step(1,3)\leftarrow,
step(2,4)\leftarrow,
step(2,5)\leftarrow,$

\hspace{9mm}$step(4,6)\leftarrow,
step(5,2)\leftarrow,
busy(3)\leftarrow,
busy(6)\leftarrow\}$.


\noindent
We would like to learn the concept $\textit{free}(X)$, meaning that the robot is free to proceed from $X$ to a non-busy, successor location. Let 


$\mathcal{E}^+=
\{\textit{free}(1), 
\textit{free}(2),
\textit{free}(5)\},$
\quad
$\mathcal{E}^- = 
\{\textit{free}(3),
\textit{free}(4),
\textit{free}(6)\}.$


\noindent
be the positive and negative examples, respectively.
Let us see how we can solve this ABA learning problem by 
our transformation rules.
We start off by applying Rote Learning and introducing the rule


$\textit{free}(X) \leftarrow X=1$\hfill(1)~~~~~

\noindent
and hence $\RR_1=\RR \cup \{(1)\}$.
Then, by Folding using the (normalised) rule 
$step(X,Y)\leftarrow X=1, Y=2$ in $\RR$, we learn

$\textit{free}(X) \leftarrow Y=2, step(X,Y)$\hfill(2)~~~~~

\noindent
and $\RR_2=\RR \cup \{
(2)\}$. By Equality Removal, we get 


$\textit{free}(X) \leftarrow step(X,Y)$\hfill(3)~~~~~

\noindent
and $\RR_2=\RR \cup \{
(3)\}$.
Rule (3) covers all positive examples, but it also 
covers the negative example $\textit{free}(4)$.
To exclude this
, we 
construct an undercutting attack to rule~(3).
We apply Assumption Introduction
to obtain $\alpha(X,Y)$ with contrary 
$\calpha(X,Y)$, and we replace rule (3) by


$\textit{free}(X) \leftarrow step(X,Y), \alpha(X,Y)$\hfill(4)~~~~~

\noindent
and $\RR_3=\RR \cup \{
(4)\}$.
The assumption $\alpha(X,Y)$ represents {\em normal} values of $(X,Y)$, for which it is legitimate 
to use default rule~(4), while $\calpha(X,Y)$
represents 
exceptions to rule (4).
Then, we add 
positive and negative examples for $\calpha(X,Y)$:


$\mathcal{E}_1^+=
\{\calpha(4,6)\}$, \quad
$\mathcal{E}_1^- = 
\{\calpha(1,2),
\calpha(2,4),
\calpha(2,5),
\calpha(5,2)\}.$

\noindent
By Rote Learning, we get 
$\RR_4=\RR \cup \{
(4),(5)\}$ with 


$\calpha(X,Y) \leftarrow X=4, Y=6$\hfill(5)~~~~~

\noindent
Finally, by Folding, using $busy(Y) \gets Y=6$, and Equality Removal, we get


$\calpha(X,Y) \leftarrow busy(Y)$\hfill(6)~~~~~

\noindent
which can be viewed as an undercutting attack to a premise of rule (4).
The final learnt set of rules is $\RR \cup \{
(4),(6)\}$, which 
indeed are a very compact and general definition of the sought concept $\textit{free}(X)$.


Now we will show that the approaches based on rebuttal encounter difficulties
in our robot example.
More specifically, we consider the approach proposed by Dimopoulos and Kakas
in~\cite{DimopoulosK95}
. Their formalism
consists of rules of the form 
%
%
$L_0 \leftarrow L_1, \ldots, L_n$
%
%
where $L_0,L_1,\ldots, L_n$ are positive or explicitly negative literals. 
These rules are understood as  default rules that admit exceptions. 
An exception is viewed as a rule whose head is the negation of some default rule it attacks, i.e., a rule of the form $\overline{L}_0 \leftarrow M_1, \ldots, M_k.$
Attacks are formalised by a priority relation between rules, 
whereby the default rule has higher priority with respect to the exception. 
As already mentioned, this type of attack is a rebuttal, in the sense that the 
default rule and the exception have opposite conclusions.
This type of rules with priorities can easily be translated into an
ABA framework, which in turn can be mapped to a normal logic program (see 
Sect.~\ref{sec:ABA}) with {\em stratified} negation, as the priority relation 
is assumed to be irreflexive and antisymmetric.

Now, from rule (3) the Dimopoulos-Kakas algorithm generates
a new learning problem for the concept $\neg \textit{free}(X)$ with 
positive and negative examples as follows:


$\mathcal{E}_1^+=
\{\neg \textit{free}(4)\}$, \quad
$\mathcal{E}_1^- = 
\{\neg \textit{free}(1),
\neg \textit{free}(2),
\neg \textit{free}(5)\}.$


\noindent
The new positive example is the exception to rule (3), and 
the new negative examples are the positive examples covered by
rule~(3), which we would like not be covered by the rules for $\neg \textit{free}(X)$.
The learning algorithm computes the rule\footnote{In fact 
the Dimopoulos-Kakas algorithm is 
just sketched and
we have conjectured what we believe to be a reasonable result of this learning step.
}


$\neg \textit{free}(X) \leftarrow step(X,Y), busy(Y)$\hfill(7)~~~~~


\noindent
which indeed covers the positive example in $\mathcal{E}_1$, but
unfortunately also covers all examples in $\mathcal{E}_1^-$.
Now, the algorithm should learn the exceptions to the exceptions,
which, however, are equal to the initial positive examples.
To avoid entering an infinite loop, the algorithm will just
enumerate these exceptions to rule~(7) by adding the set of rules
$E = \{\textit{free}(1)\leftarrow, 
\textit{free}(2)\leftarrow,
\textit{free}(5)\leftarrow\}$.
The rules in $E$ attack rule $(7)$, which in turn attacks rule $(3)$.

We 
argue that this result of the Dimopoulos-Kakas algorithm 
is due to the fact that, by rebuttal, it looks for $X$s, i.e., the variable of
the head of rule (3) that are exceptions to the rule.
In contrast, our approach also looks for exceptions that are instances of $Y$,
which does not occur in the head of the rule.

\section{A Learning Strategy}
\label{sec:strategy}

The transformation rules from Sect.~\ref{sec:transformation}
need to be guided 
to solve an ABA learning problem in a 
satisfactory way
.
For instance, 
by using our transformation rules as shown in Sect.~\ref{sec:undercutting}
we can learn a very good
solution to our robot problem
.
However, by using the transformation rules 
in a 
different way, we could also learn an ABA framework isomorphic to
the unsatisfactory Dimopoulos-Kakas solution.
In this section we present a template of a general strategy
for learning through our transformation rules (see Fig.~\ref{fig:strategy}).
This strategy is non-deterministic at several points
and different choices may lead to different final results.


\begin{figure}[ht!]
\begin{flushleft}                      \begin{small}
\textbf{Input}: ABA framework $\sABA$ (background knowledge); Training data $\mathcal{E}^+$ 
and $\mathcal{E}^-$
,
with $\mathcal{E}^+\cap \mathcal{E}^- =\emptyset$,
for concept $p_0$ to be learnt.
            

\smallskip

\noindent

\smallskip

\textbf{Repeat} the following steps \textbf{until} all examples in $\mathcal{E}^+$ are covered and all examples in $\mathcal{E}^-$ are not covered.

\smallskip

\textit{Step} 1 [\textit{Rote Learn}]
Select a predicate $p$ such that $\exists p(c)\in \mathcal{E}^+$  and, for each positive example $p(c)$ in $\mathcal{E}^+$, add the rule $p(X) \gets X=c$
to $\RR$ by Rote Learning;

\smallskip

\textit{Step} 2  [\textit{Generalise}]
For each rule in $\RR$, perform one of the following transformations:

\vspace{-3mm}

\begin{enumerate}
    \item apply Subsumption and, possibly, remove the rule from $\RR$;
    \item repeatedly apply Folding and Equality Removal until all constants are removed;
\end{enumerate}

\vspace{-2mm}


\textit{Step} 3  [\textit{Introduce Assumptions}]
Repeatedly apply Assumption Introduction as follows.

Select a rule $\rho$: $p(X) \gets Eqs, B$ (w.l.o.g. $X\subseteq vars(B)$) in $\RR$ such that
there is an argument for $p(d)\in \mathcal E^-$
with $\rho$ as top rule;
then:

\vspace{-3mm}

\begin{enumerate}
    \item select a (minimal) set $A = \{a_1(Y_1),\ldots,a_k(Y_k)\}\subseteq B$ (w.l.o.g. $a_1,\ldots,a_k$ are distinct) such that
$\exists$ disjoint sets $A^+,A^-$ of ground instances of $A$ 
such that: 

\begin{enumerate}
    \item for every example $p(e)\in \mathcal{E}^+$ covered by $\sABA$, there exist an argument for $p(e)$ {having (a ground instance of) $\rho$ as top rule} and $\{a_1(e_1),\ldots,a_k(e_k)\} \in A^+$ such that
    $a_1(e_1),\ldots,a_k(e_k)$ are children of $p(e)$;
    \item for every example $p(e)\in \mathcal{E}^-$ covered by $\sABA$ and for every argument for $p(e)$ {having (a ground instance of) $\rho$ as top rule}, there exists $\{a_1(e_1),\ldots,a_k(e_k)\} \in A^-$ such that $a_1(e_1),\ldots,a_k(e_k)$ are children of $p(e)$;
\end{enumerate}

\item Replace $\rho$ by $p(X) \gets Eqs, B, \alpha(Y_1,\ldots,Y_k)$
where $\alpha(Y_1,\ldots,Y_k)$ is a new
assumption with contrary $\chi(Y_1,\ldots,Y_k)$;

Add $\alpha(Y_1,\ldots,Y_k)$ to $\A$ and extend $\contrary{}$ by setting $\overline{\alpha(Y_1,\ldots,Y_k)}=\chi(Y_1,\ldots,Y_k)$;

\item  For every $\{a_1(e_1),\ldots,a_k(e_k)\} \in A^+$, add  $\chi(e_1,\ldots,e_k)$ to $\mathcal E^-$;

For every $\{a_1(e_1),\ldots,a_k(e_k)\} \in A^-$, add  $\chi(e_1,\ldots,e_k)$ to $\mathcal E^+$;

\end{enumerate}

\vspace{-2mm}

\textit{Step} 4: Remove from $\langle\mathcal{E}^+,\mathcal{E}^- \rangle$ all
examples for $p$.
\end{small}
\end{flushleft}   

\vspace{-3mm}

\caption{ABA Learning Strategy }
   \label{fig:strategy}

\vspace{-1mm}
\end{figure}

Let us 
briefly comment 
on the strategy.

After Step 1, all positive examples for $p$ are covered
and no negative example for $p$ is covered by 
maximally specific rules that enumerate all positive examples. 

At the end of Step 2.2, equalities between variables may be left.
After Step~2, by Property~\ref{conj:gen}, all positive examples for $p$ are still covered,
but some negative example might also be covered.

At Step 3, the condition $X\subseteq vars(B)$ on 
$p(X) \gets Eqs, B$ can always be met by adding to its body
atoms 
$true(X')$ for any missing variables $X'$.
After Step~3, by Property~\ref{conj:exception}, we get a set
of rules still covering all positive examples for $p$ and,
after Rote Learning at the 
next iteration, covers no negative example.

As it stands the ABA learning strategy might not terminate,
as
it could keep introducing new assumptions at each iteration.
However, if we assume a finite universe, 
divergence can be detected by comparing the sets of
positive/negative examples (modulo the predicate names)
generated for new assumptions
\wrt\ 
those of ``ancestor'' assumptions.
In that case we can just stop after Step 1 and
generate the trivial solution for the predicate at hand by Rote Learning.
An alternative way of enforcing termination is to allow the
application of  Assumption Introduction using 
assumptions already introduced at previous steps.
This variant of the strategy
enables learning \emph{circular debates}
as shown in the next section.

\section{Learning circular
debates 
}
\label{sec:nixon}

Existing argumentation-based approaches (notably \cite{DimopoulosK95}) are designed to learn ``stratified'' knowledge bases.
Instead, in some settings,
it is useful and natural to learn to conduct circular debates, where opinions may not be conclusively held. 
As an illustration, 
we work out a version  of the 
Nixon-diamond example \cite{nixon}. 

\smallskip

\noindent
\emph{Background Knowledge.} 
$\sABA$ with bogus assumption and 

$ \RR = \{quacker(X) \leftarrow X=a, 
republican(X) \leftarrow X=a,$

\hspace{9mm}$quacker(X) \leftarrow X=b, 
republican(X) \leftarrow X=b\}$.


\noindent
\emph{Positive Examples:}
$\mathcal{E}^+ = \{\textit{pacifist}(a)\}$, \hfill
\emph{Negative Examples:}
$ \mathcal{E}^- = \{\textit{pacifist}(b)\}.$


\noindent Note that this example can be seen as capturing a form of \emph{noise}, whereby two rows of the same table (one for $a$ and one for $b$) are characterised by exactly the same attributes (amounting to being quackers and republicans)  but have different labels (one is pacifist, the other is not). In non-monotonic reasoning terms, this requires reasoning with contradictory rules \cite{nixon}. In argumentative terms, this boils down to building circular debates. In the remainder, we show how the ABA learning strategy is able to learn these contradictory rules, in a way that circular debates can be supported.  


\smallskip

\noindent \emph{First 
iteration of the ABA learning strategy.}
At Step 1, 
we consider the positive example $\textit{pacifist}(a)$ and
by Rote Learning we add to $\RR$:


$\textit{pacifist}(X) \leftarrow X=a$   \hfill(1)~~~~~


\noindent
At Step 2, we generalise by Folding using $quacker(X) \leftarrow X=a$ in the background knowledge, and replace (1) with:

$\textit{pacifist}(X) \leftarrow quacker(X)$   \hfill(2)~~~~~

\noindent
Note that other folds 
are possible, leading to different 
ABA frameworks.

At Step 3, there is an argument for the negative example $\textit{pacifist}(b)$, because 
$quacker(X)$ holds for $X=b$.
Thus, we apply the Assumption Introduction rule to replace (2) with:

$\textit{pacifist}(X) \leftarrow quacker(X), normal\_quacker(X)$   \hfill(3)~~~~~


\noindent
with $\overline{normal\_quacker(X)}=abnormal\_quacker(X)$.
The new examples for $abnormal\_quacker$ are:

$\mathcal{E}_1^+ = \{\textit{abnormal\_quacker}(b)\},$\hfill
$ \mathcal{E}_1^- = \{\textit{abnormal\_quacker}(a)\}.$

\smallskip

\noindent
\emph{Second 
iteration of the ABA learning strategy.}
We learn the $abnormal\_quacker$ predicate. 
At Step 1, by Rote Learning we add to $\RR$:

$abnormal\_quacker(X) \leftarrow X=b$   \hfill(4)~~~~~

\noindent At Step 2, 
we fold using $republican(X) \leftarrow X=b$:

$abnormal\_quacker(X) \leftarrow republican(X)$    \hfill(5)~~~~~ 


\noindent 
There is an argument for the negative example $\textit{abnormal\_quacker}(a)$, and then, at Step 3, we
use the Assumption Introduction rule to replace (5) with:

$abnormal\_quacker(X) \leftarrow republican(X), normal\_republican(X)$   \hfill(6)~~~~~


\noindent
with $\overline{normal\_republican(X)} = abnormal\_republican(X)$.

\smallskip

\noindent \emph{Third 
iteration of the ABA learning strategy.} We learn the $abnormal\!\_republican$ predicate from the examples: 

$\mathcal{E}_2^+ = \{\textit{abnormal\_republican}(a)\}$, 
\hfill
$ \mathcal{E}_2^- = \{\textit{abnormal\_republican}(b)\}.$



\noindent
By Rote Learning we add:

$abnormal\_republican(X) \leftarrow X=a$   \hfill(7)~~~~~

\noindent We then fold using $quacker(X) \leftarrow X=a$ and we get:

$abnormal\_republican(X) \leftarrow quacker(X)$   \hfill(8)~~~~~


\noindent 
Since there is an argument for the negative example
$\textit{abnormal\_republican}(b)$, we use the Assumption Introduction rule to obtain: 

$abnormal\_republican(X) \leftarrow quacker(X), normal\_quacker(X)$   \hfill(9)~~~~~

\noindent
Note that here we use the same assumption as in (3): this is allowed by the definition of 
R5 and is useful given that rule (8) has the same body as  (2), which was replaced by rule (3) earlier on. 
%
Then, by Folding using rule (3), we get:

$abnormal\_republican(X) \leftarrow \textit{pacifist}(X)$   \hfill(10)~~~~~

\noindent This leads to an ABA framework with final set of learnt rules 
\{(3), (6), (10)\}, encompassing a circular debate whereby an argument for 
being pacifist is attacked by one for being non-pacifist
(and vice versa).
This ABA framework has (among others) the following stable extension
(corresponding to the choice,  as the  accepted set of assumptions,  of $\{normal\_quacker(a), normal\_republican(b)\}$):


$\{\argu{\{normal\_quacker(a)\}}{\textit{pacifist}(a)},$
   
   \hspace{1.7mm}$\argu{\{normal\_quacker(a)\}}{normal\_quacker(a)}, $
   
\hspace{1.7mm}$\argu{\{\textit{normal\_quacker}(a)\}}{abnormal\_republican(a)},$

\hspace{1.7mm}$\argu{\{normal\_republican(b)\}}{abnormal\_quacker(b)},$

\hspace{1.7mm}$
\argu{\{normal\_republican(b)\}}{normal\_republican(b)}$,

\hspace{1.7mm}$\argu{\emptyset}{quacker(a)}, \argu{\emptyset}{republican(a)}, \argu{\emptyset}{quacker(b)}, \argu{\emptyset}{republican(b)}
\}$.

\noindent
Here, $a$ is pacifist and $b$ is not, thus achieving the goal of ABA learning under credulous reasoning.   Note that there are three other stable extensions of the resulting ABA framework (one where $b$ is pacifist and $a$ is not, one where both are pacifist and one where neither is), and thus sceptical reasoning would not~work.

\section{Conclusions}

We have presented a novel approach to learning ABA frameworks from positive and negative examples, using 
background knowledge. 
A notable feature of our 
method
is that it is able to learn exceptions to general rules 
that can be interpreted as {undercutting attacks},
besides the more traditional {rebuttal attacks}.
Our learning technique is based on 
transformation rules, borrowing a well established approach from logic program transformation.
We have presented a general strategy that applies these transformation 
rules in a suitable order 
and we have shown through some examples
that 
we are
able, on one hand, to reconstruct other logic-based learning approaches that learn stratified rules
and, on the other hand, 
to 
also learn rules 
enacting
non-stratified, circular debates.

Our 
approach can be expanded 
in  several directions.
First of all, we need to study conditions guaranteeing
that our transformation rules always derive a correct 
solution 
(see Properties~\ref{conj:gen}--\ref{conj:exception}).
To do that we may build upon results available in the field of logic program transformation.
At present, our learning strategy is a template relying
on several non-deterministic choices
that need to be realised to obtain a
concrete algorithm.
Then, we 
plan to conduct experiments on existing
benchmarks 
in comparison 
with other systems.
In particular, for 
comparison,
we 
plan to use FOLD~\cite{ShakerinSG17},
which learns stratified 
normal 
logic programs
, 
and ILASP~\cite{LawRB14}, which learns Answer Set Programs.
We also plan to explore whether solutions to our ABA learning problems are better suited, than existing approaches to ILP, to provide explanations, leveraging on several existing approaches to argumentative XAI~\cite{survey}.   

\vspace*{-0.1cm}

\subsubsection*{Acknowledgements.} We thank the anonymous reviewers for 
useful comments.
We also 
thank Mark Law for 
advice on 
the ILASP system.
F. Toni was partially funded by the European Research Council (ERC) under
the European Union’s Horizon 2020 research and innovation programme (grant
agreement No. 101020934) and 
by 
J.P. Morgan and  the
Royal Academy of Engineering under the Research Chairs
and Senior Research Fellowships scheme. 
M. Proietti is a member of the INdAM-GNCS research group.

\bibliographystyle{alpha}
\bibliography{main}

\newcommand{\etalchar}[1]{$^{#1}$}
\begin{thebibliography}{CRA{\etalchar{+}}21}

\bibitem[ABW88]{Ap&88}
K.R. Apt, H.A. Blair, and A.~Walker.
\newblock Towards a theory of declarative knowledge.
\newblock In {\em Foundations of Deductive Databases and Logic Programming.},
  pages 89--148. Morgan Kaufmann, 1988.

\bibitem[AD95]{ArD95}
C.~Aravindan and P.M. Dung.
\newblock On the correctness of unfold/fold transformation of normal and
  extended logic programs.
\newblock {\em J. Log. Program.}, 24(3):201--217, 1995.

\bibitem[BDKT97]{ABA}
A.~Bondarenko, P.M. Dung, R.A. Kowalski, and F.~Toni.
\newblock An abstract, argumentation-theoretic approach to default reasoning.
\newblock {\em Artif. Intell.}, 93:63--101, 1997.

\bibitem[CFST17]{ABAhandbook}
K.~Cyras, X.~Fan, C.~Schulz, and F.~Toni.
\newblock Assumption-based argumentation: Disputes, explanations, preferences.
\newblock {\em {FLAP}}, 4(8), 2017.

\bibitem[CRA{\etalchar{+}}21]{survey}
K.~Cyras, A.~Rago, E.~Albini, P.~Baroni, and F.~Toni.
\newblock Argumentative {XAI:} {A} survey.
\newblock In {\em IJCAI}, pages 4392--4399, 2021.

\bibitem[CSCT20]{dear}
O.~Cocarascu, A.~Stylianou, K.~Cyras, and F.~Toni.
\newblock Data-empowered argumentation for dialectically explainable
  predictions.
\newblock In {\em {ECAI}}, pages 2449--2456, 2020.

\bibitem[CT16]{ML-argsurvey}
O.~Cocarascu and F.~Toni.
\newblock Argumentation for machine learning: {A} survey.
\newblock In {\em {COMMA}}, pages 219--230, 2016.

\bibitem[DK95]{DimopoulosK95}
Y.~Dimopoulos and A.~C. Kakas.
\newblock Learning non-monotonic logic programs: Learning exceptions.
\newblock In {\em ECML}, LNCS 912, pages 122--137. Springer, 1995.

\bibitem[DKT09]{ABAbook}
P.M. Dung, R.A. Kowalski, and F.~Toni.
\newblock Assumption-based argumentation.
\newblock In {\em Arg. in AI}, pages 199--218. Springer, 2009.

\bibitem[Dun95]{Dung_95}
P.M. Dung.
\newblock {On the Acceptability of Arguments and its Fundamental Role in
  Nonmonotonic Reasoning, Logic Programming and n-Person Games}.
\newblock {\em Artif. Intell.}, 77(2):321--358, 1995.

\bibitem[GL88]{SM}
M.~Gelfond and V.~Lifschitz.
\newblock The stable model semantics for logic programming.
\newblock In {\em ICLP}, pages 1070--1080. {MIT} Press, 1988.

\bibitem[IH00]{InoueH00}
K.~Inoue and H.~Haneda.
\newblock Learning abductive and nonmonotonic logic programs.
\newblock In {\em Abduction and Induction: Essays on their Relation and
  Integration}, pages 213--231. Kluwer Academic, 2000.

\bibitem[IK97]{InoueK97}
K.~Inoue and Y.~Kudoh.
\newblock Learning extended logic programs.
\newblock In {\em IJCAI}, pages 176--181. Morgan Kaufmann, 1997.

\bibitem[LRB14]{LawRB14}
M.~Law, A.~Russo, and K.~Broda.
\newblock Inductive learning of answer set programs.
\newblock In {\em {JELIA}}, LNCS 8761, pages 311--325. Springer, 2014.

\bibitem[Mug95]{Muggleton1995}
S.~Muggleton.
\newblock Inverse entailment and {P}rogol.
\newblock {\em New generation computing}, 13(3-4):245--286, 1995.

\bibitem[PP94]{PeP94}
A.~Pettorossi and M.~Proietti.
\newblock Transformation of logic programs: Foundations and techniques.
\newblock {\em J. Log. Program.}, 19/20:261--320, 1994.

\bibitem[Ray09]{Ray09}
O.~Ray.
\newblock Nonmonotonic abductive inductive learning.
\newblock {\em J. Appl. Log.}, 7(3):329--340, 2009.

\bibitem[RC81]{nixon}
R.~Reiter and G.~Criscuolo.
\newblock On interacting defaults.
\newblock In {\em IJCAI}, pages 270--276. William Kaufmann, 1981.

\bibitem[Sak00]{Sakama00}
C.~Sakama.
\newblock Inverse entailment in nonmonotonic logic programs.
\newblock In {\em ILP}, LNCS 1866, pages 209--224. Springer, 2000.

\bibitem[Sak05]{Sakama05}
C.~Sakama.
\newblock Induction from answer sets in nonmonotonic logic programs.
\newblock {\em {ACM} Trans. Comput. Log.}, 6(2):203--231, 2005.

\bibitem[Sek91]{Sek91}
H.~Seki.
\newblock Unfold/fold trans\-form\-ation of stratified pro\-grams.
\newblock {\em Theoretical Computer Science}, 86:107--139, 1991.

\bibitem[SI09]{SakamaI09}
C.~Sakama and K.~Inoue.
\newblock Brave induction: {A} logical framework for learning from incomplete
  information.
\newblock {\em Mach. Learn.}, 76(1):3--35, 2009.

\bibitem[SSG17]{ShakerinSG17}
F.~Shakerin, E.~Salazar, and G.~Gupta.
\newblock A new algorithm to automate inductive learning of default theories.
\newblock {\em TPLP}, 17(5-6):1010--1026, 2017.

\bibitem[TK96]{ToK96}
F.~Toni and R.~Kowalski.
\newblock An argumentation-theoretic approach to logic program transformation.
\newblock In {\em LOPSTR}, LNCS 1048, pages 61--75, 1996.

\bibitem[Ton14]{ABAtutorial}
F.~Toni.
\newblock A tutorial on assumption-based argumentation.
\newblock {\em Arg. \& Comp}, 5(1):89--117, 2014.

\end{thebibliography}
\end{document}